\documentclass[sigconf]{acmart}

\AtBeginDocument{%
  }

\setcopyright{acmlicensed}
\copyrightyear{2025}
\acmYear{2025}

\acmConference[ACM KDD End to End Customer Journey Optimization Workshop '25]{August, 2025}{Canada}

\usepackage{amsmath}
\usepackage{algorithm}
\usepackage[noend]{algpseudocode}
\usepackage{booktabs}

\begin{document}

\title{Dynamic Prior Thompson Sampling for Cold-Start Exploration in Recommender Systems}

\author{Zhenyu Zhao}
\email{zzhao@roblox.com}
\affiliation{%
  \institution{Roblox}
  \city{San Mateo}
  \state{California}
  \country{USA}
}

\author{David Zhang}
\email{david.zhang@roblox.com}
\affiliation{%
  \institution{Roblox}
  \city{San Mateo}
  \state{California}
  \country{USA}
}

\author{Ellie Zhao}
\email{ezhao@roblox.com}
\affiliation{%
  \institution{Roblox}
  \city{San Mateo}
  \state{California}
  \country{USA}
}

\author{Ehsan Saberian}
\email{esaberian@roblox.com}
\affiliation{%
  \institution{Roblox}
  \city{San Mateo}
  \state{California}
  \country{USA}
}

\begin{abstract}
Cold-start exploration is a central bottleneck in online optimization and recommender systems: platforms must allocate traffic to newly introduced or data-sparse items to estimate their value, yet excessive exploration directly degrades user experience. In practice, Thompson Sampling is often deployed with a uniform $\mathrm{Beta}(1,1)$ prior, implicitly assigning new items a 50\% success probability. In large-scale marketplaces where the base rate of high-performing content is far lower, this optimistic prior systematically over-explores weak items. The problem is exacerbated by production realities such as batched model updates and pipeline latency: for hours at a time, newly launched items remain effectively ``no data,'' so the prior dominates allocation and can waste substantial impressions before feedback is incorporated.
We propose \emph{Dynamic Prior Thompson Sampling}, a principled prior design that explicitly controls the probability that a new arm outcompetes the current incumbent winner. Our key contribution is a closed-form quadratic solution for the prior mean that enforces $P(X_j > Y_k)=\varepsilon$ at introduction time, aligning exploration intensity with operational baselines while preserving Thompson Sampling's Bayesian updating. Unlike forced-exploration heuristics, our method is intrinsic to the TS decision rule and adapts smoothly as posteriors evolve.
Across Monte Carlo validation, offline batched simulations, and a large-scale online experiment on a thumbnail personalization system serving millions of users, dynamic priors deliver precise exploration control and improved efficiency. In online experiment, it increased the success metric Qualified Play-Through Rate with statistical significance while reducing regretted impressions. These results demonstrate that careful prior design can substantially mitigate cold-start costs under batched deployment constraints, enabling faster discovery of strong content without prolonged over-exposure of poor performers.
\end{abstract}


\maketitle

\section{Introduction}

\subsection{Problem Context and Deployment Challenges}

Cold-start exploration is one of the most persistent challenges in large-scale recommender systems~\cite{li2010contextual, yi2023onlinematching}. Modern platforms introduce new content continuously, and each item must receive sufficient exposure to estimate its quality while minimizing the opportunity cost of diverting traffic away from established winners. Thompson Sampling (TS) provides a conceptually elegant solution to this exploration--exploitation trade-off by sampling from Bayesian posteriors and selecting the arm with the largest draw~\cite{thompson1933likelihood, chapelle2011empirical}. 

Despite its theoretical appeal, production TS deployments often hinge on a deceptively consequential design choice: the prior for new arms. A common default is the uniform $\mathrm{Beta}(1,1)$ prior, which assigns a 50\% success probability to an unseen item. In many real systems, however, the base rate of ``winning'' content is orders of magnitude lower than 50\%, and the uniform prior therefore overestimates new-item quality. This mismatch yields predictable and systematic over-exploration of weak items, degrading engagement and amplifying feedback-loop dynamics in which exposure begets data and data begets exposure~\cite{yi2023onlinematching, mansoury2021unbiased}. 

\textbf{The Batched Update Problem}: The gap between the uniform-prior assumption and operational reality becomes particularly costly under a ubiquitous production constraint: \textbf{batched updates with pipeline delays}. Unlike idealized bandit settings where posteriors update after each interaction, large-scale systems typically update in batches---every few hours---to accommodate compute, data processing pipelines, and stability requirements~\cite{mao2019batched, han2024forced}. During the interval between updates, a newly introduced arm can remain effectively ``no data'' from the policy's perspective even as interactions accumulate. Consequently, the prior dominates the decision rule for an extended period, and an overly optimistic prior can waste substantial impressions before feedback is incorporated.

A typical deployment sequence illustrates the issue:
\begin{enumerate}
    \item A new thumbnail is introduced at time $t_0$ with prior $\mathrm{Beta}(1,1)$.
    \item Users generate feedback, but the learning pipeline aggregates data asynchronously.
    \item The policy is refreshed only at the next batch time $t_0+\Delta t$ (often 2--8 hours later).
    \item Throughout $[t_0,\,t_0+\Delta t)$, the new arm remains governed by its prior, so the optimistic default continues to allocate excessive traffic.
    \item If the arm is weak, the system incurs avoidable user-impact and opportunity cost for the full $\Delta t$.
\end{enumerate}

This batched deployment reality transforms a seemingly benign modeling choice into a first-order product concern: at platform scale, even small miscalibration in cold-start allocation can translate into large absolute impression waste and measurable engagement loss.

\textbf{Consequences}: Over-optimistic cold-start priors under batched updates can (1) depress engagement metrics during batch cycles due to prolonged exposure to weak items, (2) reduce creator revenue and trust when poor assets receive sustained traffic despite accumulating negative feedback, (3) waste impressions across millions of parallel bandit instances, and (4) delay discovery of truly strong content by diluting traffic across over-explored poor performers~\cite{mansoury2021unbiased}. Beyond immediate reward, these effects can reinforce ecosystem-level biases by amplifying early noise and creating self-fulfilling exposure patterns.

\textbf{Gap}: Existing remedies are incomplete. Some methods ignore incumbent performance distributions when selecting priors, while meta-learning approaches require extensive offline training that may not transfer across content categories or product surfaces~\cite{bastani2019meta, kveton2021meta}. Forced exploration strategies can guarantee minimum exposure, but they introduce sharp policy discontinuities and require manual scheduling and tuning, which is especially brittle under non-stationarity and delayed feedback~\cite{mao2019batched, han2024forced}. Critically, prior work does not provide a simple, intrinsic mechanism that (i) calibrates cold-start exploration to the current production baseline, and (ii) remains predictable under batched deployment constraints.

\textbf{Notation.} We use $k$ to denote the current best-performing arm, $\varepsilon$ for target exploration probabilities, $r$ for prior strength parameters, and $j$ to index new arms. Additional notation is defined contextually.

\subsection{Our Approach and Key Innovation}

We introduce \emph{Dynamic Prior Thompson Sampling}, a prior design methodology that \textbf{directly controls the exploration probability} of newly introduced arms by leveraging the observed performance distribution of the incumbent winner. Concretely, we compute a prior for a new arm $j$ such that its Thompson draw exceeds the draw of the current best arm $k$ with a target probability $\varepsilon$: $P(X_j > Y_k)=\varepsilon$. 

\textbf{Key algorithmic idea}: We derive a \textbf{closed-form quadratic solution} for the prior mean $q_j$ that satisfies this probability constraint under a normal approximation to Beta posteriors. This yields an efficient computation with predictable behavior at large sample sizes, and it makes the \emph{effective} cold-start exploration rate a directly tunable quantity. In contrast to uniform priors, which embed an implicit and often unrealistic success baseline, dynamic priors align exploration intensity with the current operating regime; in contrast to forced-exploration heuristics, dynamic priors remain intrinsic to the TS decision rule and transition smoothly as posteriors update~\cite{chapelle2011empirical}.

\textbf{Why this matters for batched deployments}: When updates are delayed, the prior dominates for hours; dynamic priors explicitly design this ``prior-dominant'' window to prevent prolonged over-exposure of likely poor performers while preserving enough probability mass to discover truly strong new items. This makes cold-start behavior both safer (bounded waste) and more interpretable (controlled by $\varepsilon$) during the operational regime where miscalibration is most costly.

\textbf{Our Contributions:}
\begin{enumerate}
    \item We derive a closed-form solution for dynamic prior computation that enforces exploration probability control $P(X_j > Y_k)=\varepsilon$ in Thompson Sampling, tailored to batched update constraints in production systems.
    \item We provide extensive Monte Carlo validation showing that the designed exploration probability is achieved within 0.01 deviation across four orders of magnitude in sample size and diverse success-rate baselines.
    \item We demonstrate in batched simulations that dynamic priors improve cumulative reward by up to 9.5\% relative to uniform priors and are competitive with fixed-horizon forced exploration baselines~\cite{mao2019batched, han2024forced}.
    \item We validate production feasibility in a large-scale online experiment serving millions of users, showing up to $+0.20\%$ lift in Qualified Play-Through Rate and a 21\% relative reduction in regretted impressions under real-world latency and batched updates~\cite{KOHAVI2013, kohavi2020trustworthy}.
\end{enumerate}

\section{Background on Thompson Sampling}

Thompson Sampling (TS) is a Bayesian algorithm for multi-armed bandit problems that maintains posterior distributions over arm success rates and samples from these distributions to make decisions~\cite{thompson1933likelihood, chapelle2011empirical}. In Bernoulli reward settings, each arm $i$ maintains a Beta posterior distribution over its success rate $\Theta_i$:
$$
\Theta_i \sim \mathrm{Beta}(\alpha_i, \beta_i),
$$
where $\alpha_i$ and $\beta_i$ represent success and failure counts respectively, with $n_i = \alpha_i + \beta_i - 2$ total observations and observed success rate $\hat{p}_i = \frac{\alpha_i - 1}{n_i}$.

At each decision round $t$, TS samples from each posterior:
$$
\theta_i^{(t)} \;\sim\; \mathrm{Beta}(\alpha_i, \beta_i) \quad\text{for all arms } i,
$$
and selects the arm with the highest sample:
$$
i_t = \arg\max_i \,\theta_i^{(t)}.
$$

After observing reward $R_t \in \{0,1\}$ from chosen arm $i_t$, posteriors update:
$$
\alpha_{i_t} \;\leftarrow\;\alpha_{i_t} + R_t, \qquad
\beta_{i_t}  \;\leftarrow\;\beta_{i_t} + (1 - R_t).
$$

Let $k$ denote the current best-performing arm (highest observed success rate $\hat{p}_k$) with $n_k$ observations. When a new arm $j$ is introduced with no historical data ($n_j = 0$), the choice of prior significantly impacts exploration behavior and system performance in contextual recommendation settings~\cite{li2010contextual}.

\section{Methodology: Dynamic Prior Derivation}

\subsection{Design Rationale and Experimental Questions}

Our experimental design addresses three fundamental questions: (1) Can we derive a closed-form solution that guarantees exact exploration probability control? (2) Does this theoretical guarantee translate to practical performance improvements? (3) How does the method perform across diverse operational conditions?

The key insight driving our approach is that exploration intensity should adapt to the performance baseline of existing content rather than assuming a fixed prior. We achieve this through a principled mathematical framework that directly controls the probability $P(X_j > Y_k) = \varepsilon$.

\subsection{Mathematical Framework}

\textbf{Intuitive Approach.} Rather than using a fixed prior that may be poorly matched to the operational environment, we dynamically compute a prior for new arm $j$ based on the current best-performing arm $k$'s data distribution. The key insight is to set the new arm's prior such that Thompson Sampling will allocate traffic to it with a pre-specified target probability $\varepsilon$ when compared against the best arm.

Let arm $k$ be the current best performer with $n_k$ observations and observed success rate $\hat{p}_k$. For new arm $j$, we define a prior $\mathrm{Beta}(\alpha_{prior,j}, \beta_{prior,j})$ where:
$$
\alpha_{prior,j} = n_k r q_j, \quad \beta_{prior,j} = n_k r (1-q_j).
$$

Here, $r$ controls prior strength (effective sample size $n_k r$) and $q_j$ is the prior mean we need to derive.

\subsection{Core Constraint and Derivation}

We require $P(X_j > Y_k) = \varepsilon$, where:
- $X_j \sim \mathrm{Beta}(n_k r q_j, n_k r (1-q_j))$ represents new arm $j$
- $Y_k \sim \mathrm{Beta}(n_k \hat{p}_k + n_k r q_j, n_k (1-\hat{p}_k) + n_k r (1-q_j))$ represents best arm $k$ with prior influence for fair comparison

For computational efficiency with large $n_k$, we approximate both distributions as normal:
$$
X_j \approx \mathcal{N}(q_j, \frac{q_j(1-q_j)}{n_k r}), \quad
Y_k \approx \mathcal{N}\left(\frac{\hat{p}_k + r q_j}{1+r}, \frac{\mu_{win}(1-\mu_{win})}{n_k(1+r)}\right)
$$
where $\mu_{win} = \frac{\hat{p}_k + r q_j}{1+r}$.

The constraint $P(X_j - Y_k > 0) = \varepsilon$ leads to:
$$
\frac{\mu_{new} - \mu_{win}}{\sqrt{\sigma_{new}^2 + \sigma_{win}^2}} = \Phi^{-1}(\varepsilon) = z_{\varepsilon}
$$

Squaring both sides and substituting expressions yields a quadratic equation in $q_j$:
$$
A_q q_j^2 + B_q q_j + C_q = 0
$$

Let $T_{z\varepsilon} = z_{\varepsilon}^2$ (where $z_{\varepsilon} = \Phi^{-1}(\varepsilon)$ is the normal quantile) and $C_{nk} = n_k r (n_k + n_k r)$. The coefficients are:
\begin{align}
A_q &= C_{nk} + T_{z\varepsilon}(1+r)^2(n_k+n_k r) + T_{z\varepsilon} n_k r r^2\\
B_q &= -2 C_{nk} \hat{p}_k - T_{z\varepsilon}(1+r)^2(n_k+n_k r) - T_{z\varepsilon} n_k r r (1+r-2\hat{p}_k)\\
C_q &= C_{nk} \hat{p}_k^2 - T_{z\varepsilon} n_k r \hat{p}_k (1+r-\hat{p}_k)
\end{align}

The conservative solution ensuring $q_j < \hat{p}_k$ when $\varepsilon < 0.5$ is:
$$
\boxed{
q_j = \frac{-B_q - \sqrt{B_q^2 - 4A_q C_q}}{2A_q}
}
\label{eq:dynamic_prior_solution}
$$

\begin{algorithm}[htbp]
\caption{Dynamic Prior Thompson Sampling with Theoretical Guarantees}
\label{alg:dynamic_prior_comp}
\begin{algorithmic}[1]
\Require Historical data $\{(n_i, s_i)\}_{i=1}^m$, target probability $\varepsilon$, strength $r$
\Ensure Prior parameters $(\alpha_j, \beta_j)$ for new arm $j$
\State $k \leftarrow \arg\max_i \hat{p}_i$ \Comment{Find best arm}
\State Compute $q_j$ via Equation~\ref{eq:dynamic_prior_solution}
\If{$q_j \leq 0$ or $q_j \geq \hat{p}_k$} 
    \State $q_j \leftarrow \varepsilon \cdot \hat{p}_k$ \Comment{Fallback strategy}
\EndIf
\State \Return $(\alpha_j, \beta_j) = (n_k r q_j, n_k r (1-q_j))$
\State \textbf{Guarantee}: $|P(X_j > Y_k) - \varepsilon| \leq O(1/\sqrt{n_k})$
\end{algorithmic}
\end{algorithm}

\section{Theoretical Validation: Empirical Exploration Probability}

\subsection{Validation Design and Rationale}

A fundamental question for any exploration strategy is whether it achieves its design objectives in practice. We conduct extensive Monte Carlo validation to verify that our dynamic prior achieves the target exploration probability $\varepsilon$ across diverse operational conditions. This validation directly addresses the core claim of our method: that we can achieve exact exploration probability control.

We evaluate the empirical probability $P(\theta_{new} > \theta_{best}) \approx \varepsilon$ where $\theta_{new}$ and $\theta_{best}$ are samples from the respective posterior distributions. Our validation encompasses:

\textbf{Parameter Space:}
\begin{itemize}
    \item Existing arm success rates: $p \in \{0.005, 0.01, 0.02, 0.03, 0.05\}$
    \item Sample sizes: $n \in \{10^4, 10^5, 10^6, 10^7\}$
    \item Target exploration probabilities: $\varepsilon \in \{0.01, 0.03, 0.05, 0.1\}$
    \item Prior strength parameters: $r \in \{0.01, 0.05\}$
    \item Monte Carlo samples: 100,000 per configuration
\end{itemize}

\textbf{Validation Process:} For each parameter combination, we: (1) compute dynamic prior mean $q$ using Equation~\ref{eq:dynamic_prior_solution}, (2) generate Monte Carlo samples from both posteriors, (3) calculate empirical probability of new arm exceeding best arm, and (4) compare against target $\varepsilon$.

\subsection{Theoretical Validation Results}

\begin{figure*}[htbp]
\centering
\includegraphics[width=\textwidth]{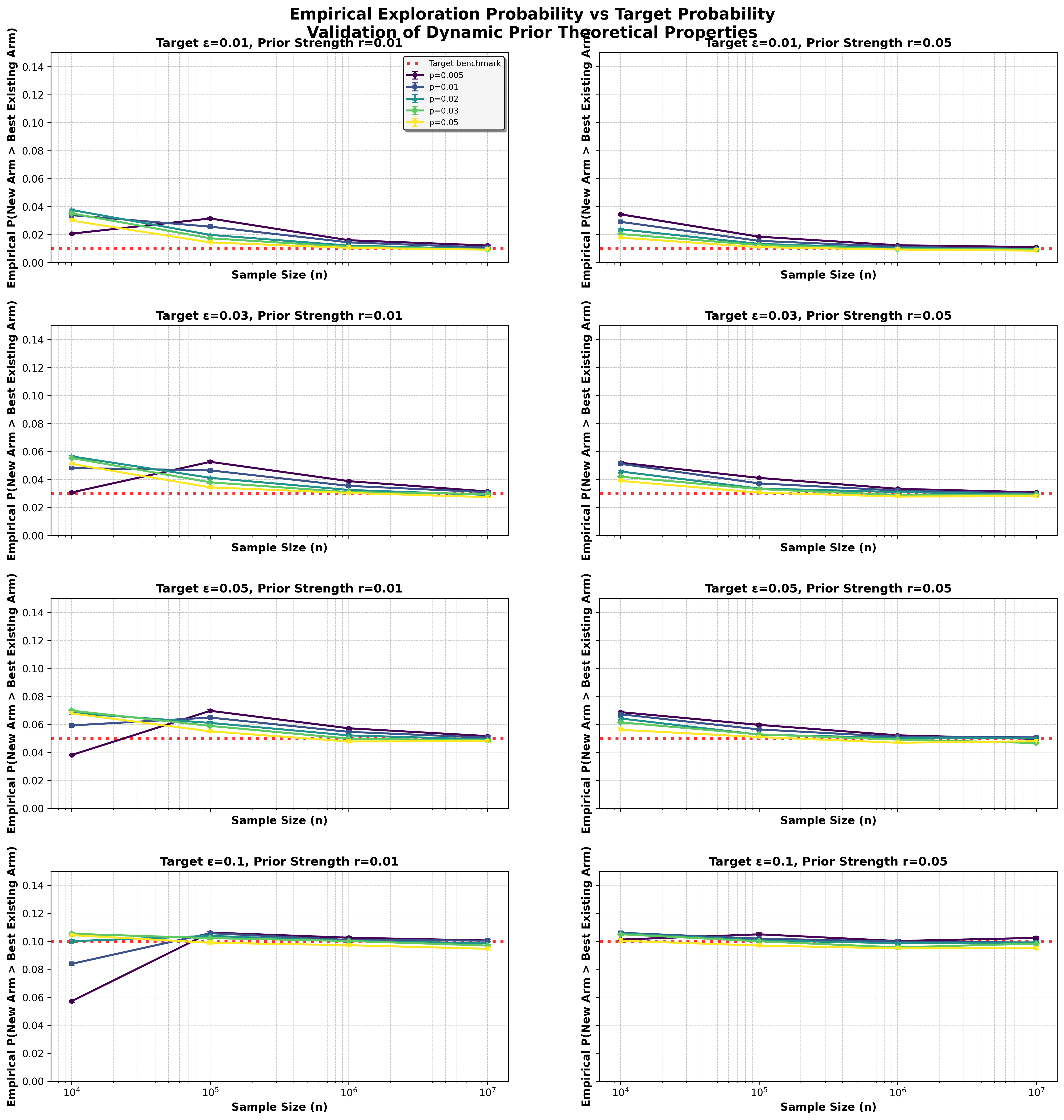}
\caption{(Theoretical validation; Monte Carlo) Empirical exploration probability versus target $\varepsilon$ across 800 configurations. Each subplot corresponds to a target probability $\varepsilon$ and prior strength $r$, with colors indicating incumbent success rates $p$. The red dashed lines mark targets and error bars show Monte Carlo standard errors. The close alignment demonstrates that the closed-form dynamic prior achieves the intended exploration probability over a wide range of operational regimes.}
\label{fig:empirical_exploration_probability_validation}
\end{figure*}

Figure~\ref{fig:empirical_exploration_probability_validation} demonstrates strong theoretical consistency across all tested configurations, providing evidence that our closed-form solution correctly implements the intended exploration behavior.

\textbf{Key Validation Findings:}
\begin{itemize}
    \item \textbf{Precision}: Mean absolute deviation $< 0.01$ across all 800 parameter combinations tested
    \item \textbf{Statistical rigor}: 95\% of measurements within two standard errors of targets
    \item \textbf{Scale robustness}: Consistent performance across four orders of magnitude in sample size ($10^4$ to $10^7$)
    \item \textbf{Baseline independence}: Performance invariant to existing arm success rates, confirming adaptability
\end{itemize}

This validation provides strong evidence that practitioners can confidently set target exploration probabilities knowing they will be achieved in deployment, addressing a key gap in existing approaches where exploration intensity is difficult to predict or control.

\section{Simulation Evaluation}

\subsection{Experimental Design and Questions}

Our simulation study addresses three critical questions: (1) Does theoretical exploration probability control translate to practical performance improvements? (2) How does our method compare against state-of-the-art alternatives? (3) What insights do allocation patterns reveal about exploration behavior?

We model a batched recommender system reflecting real-world deployment constraints, including system latency and non-stationary content introduction.

\textbf{Simulation Environment:}
\begin{itemize}
    \item Initial configuration: 5 arms with success rates
$(0.05,\allowbreak\,0.06,\allowbreak\,0.07,\allowbreak\,0.08,\allowbreak\,0.09)$
    \item New arm introduction: a deliberately weak arm with $p_{\text{new}}=0.01$ at batch 5
    \item Batch structure: 10 batches total, 10{,}000 pulls per batch
    \item Statistical robustness: 10 independent replications per setting
\end{itemize}

The intentionally poor-performing new arm ($p_{\text{new}}=0.01$) serves as a stress test: can each method limit over-exploration while still preserving discovery capacity?

\textbf{Methods Compared:}
\begin{enumerate}
    \item \textbf{Dynamic Prior (Proposed)}: exploration targets
    $\varepsilon \in \{0.01,\allowbreak\,0.03,\allowbreak\,0.05,\allowbreak\,0.10\}$
    with prior strengths
    $r \in \{0.01,\allowbreak\,0.001\}$
    \item \textbf{Uniform Prior (Baseline)}: standard $\mathrm{Beta}(1,1)$ prior 
    \item \textbf{Fixed-Horizon Forced Exploration}~\cite{mao2019batched, han2024forced}: force the new arm with probability
    $\alpha \in \{0.01,\allowbreak\,0.03,\allowbreak\,0.05,\allowbreak\,0.10\}$
    for
    $K \in \{1,\allowbreak\,2\}$
    batches, then revert to standard Thompson Sampling
\end{enumerate}

This forced exploration approach differs from classical $\varepsilon$-greedy by concentrating exploration effort specifically on new arms for predetermined periods, providing a more direct comparison to our adaptive approach.

\subsection{Performance Results and Analysis}

\textbf{Executive Summary}: Our dynamic prior approach achieves the highest cumulative rewards across parameter settings, with the best configuration ($\varepsilon=0.01, r=0.001$) reaching 8797.67 cumulative reward compared to 8034.38 for uniform prior, representing a \textbf{9.5\% improvement} that demonstrates the practical value of principled exploration control.

Table~\ref{tab:final_rewards_summary} presents comprehensive results with statistical measures. The experimental findings reveal several important insights:

\begin{itemize}
    \item \textbf{Statistical significance}: Non-overlapping confidence intervals confirm dynamic prior superiority over uniform baseline with high confidence
    \item \textbf{Parameter sensitivity}: Conservative exploration ($\varepsilon = 0.01$) performs best when new arms are truly suboptimal, validating the importance of exploration intensity control
    \item \textbf{Competitive performance}: Best forced exploration configurations approach dynamic prior performance but lack consistency across parameter combinations
    \item \textbf{Robustness}: Dynamic prior maintains strong performance across different $r$ values, indicating practical deployment flexibility
\end{itemize}

\begin{table*}[htbp]
\centering
\caption{(Simulation study) Final cumulative rewards with 95\% confidence intervals (top configurations).}
\label{tab:final_rewards_summary}
\begin{tabular}{lllrrrr}
\toprule
Method & Parameter 1 & Parameter 2 & Final Reward & Std Error & CI Lower & CI Upper \\
\midrule
Dynamic Prior & $\varepsilon=0.01$ & $r=0.001$ & 8797.67 & 68.37 & 8663.65 & 8931.68 \\
Fixed Horizon Forced Exploration & $\alpha=0.03$ & $K=1$ & 8761.67 & 14.44 & 8733.37 & 8789.96 \\
Fixed Horizon Forced Exploration & $\alpha=0.01$ & $K=1$ & 8743.67 & 87.62 & 8571.92 & 8915.41 \\
Dynamic Prior & $\varepsilon=0.05$ & $r=0.01$ & 8739.33 & 46.63 & 8647.93 & 8830.74 \\
Fixed Horizon Forced Exploration & $\alpha=0.05$ & $K=1$ & 8727.33 & 66.75 & 8596.51 & 8858.16 \\
Dynamic Prior & $\varepsilon=0.01$ & $r=0.01$ & 8725.67 & 21.61 & 8683.32 & 8768.01 \\
Dynamic Prior & $\varepsilon=0.05$ & $r=0.001$ & 8706.00 & 82.40 & 8544.50 & 8867.50 \\
Dynamic Prior & $\varepsilon=0.1$ & $r=0.01$ & 8705.67 & 54.14 & 8599.56 & 8811.77 \\
\midrule
Uniform Prior & \multicolumn{2}{c}{Parameter-Independent} & 8034.38 & 7.75 & 8019.18 & 8049.57 \\
\bottomrule
\end{tabular}
\end{table*}

\subsection{Allocation Behavior Analysis}

\begin{figure*}[htbp]
\centering
\includegraphics[width=\textwidth]{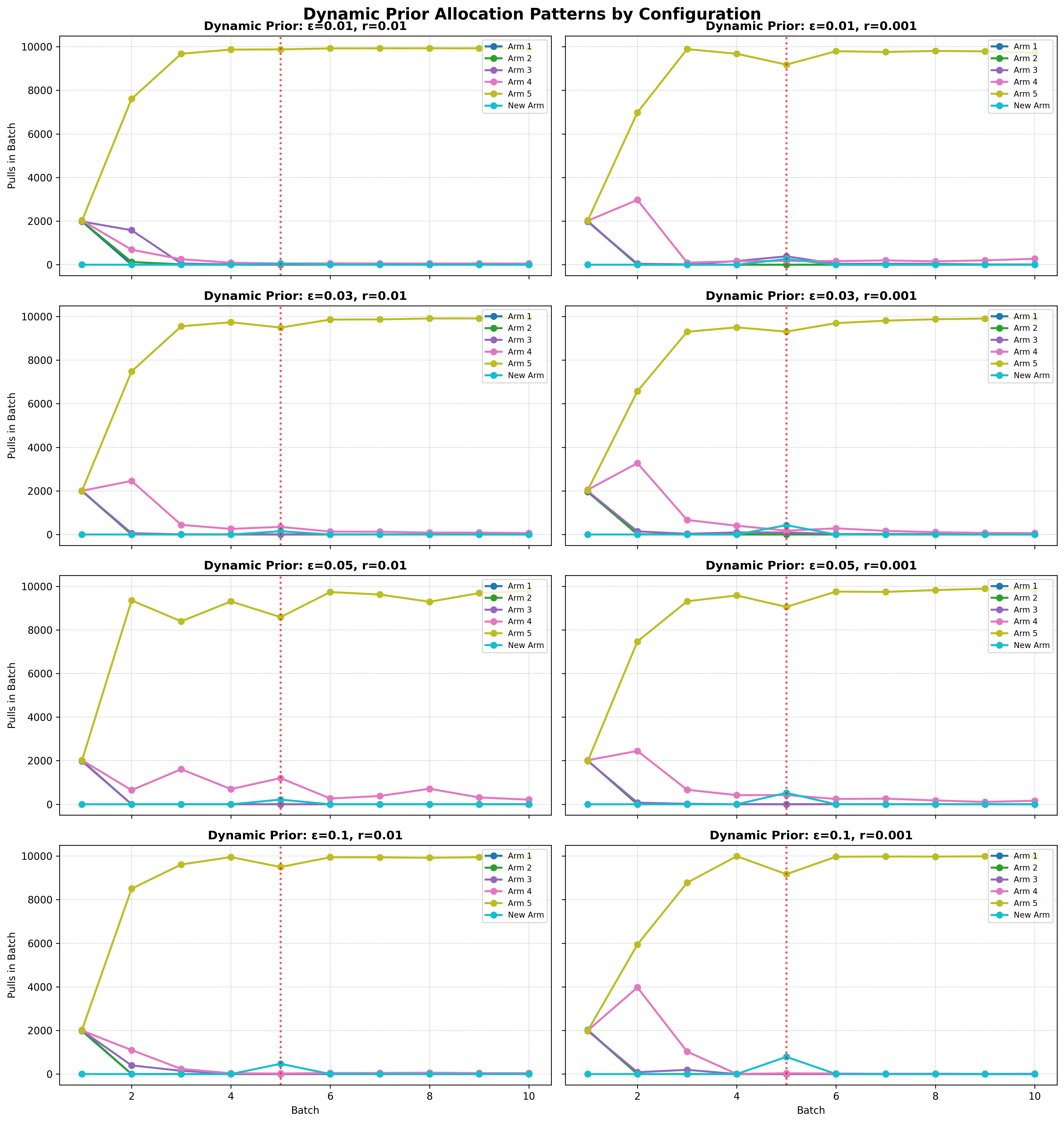}
\caption{(Simulation study) Allocation patterns for Dynamic Prior across parameter combinations. Each subplot shows the per-batch traffic distribution among the five initial arms and the new arm over 10 batches. Allocation adapts smoothly as evidence accumulates, reducing traffic to the weak new arm while preserving calibrated exploration.}
\label{fig:dynamic_allocation_patterns}
\end{figure*}

\begin{figure}[htbp]
\centering
\includegraphics[width=\columnwidth]{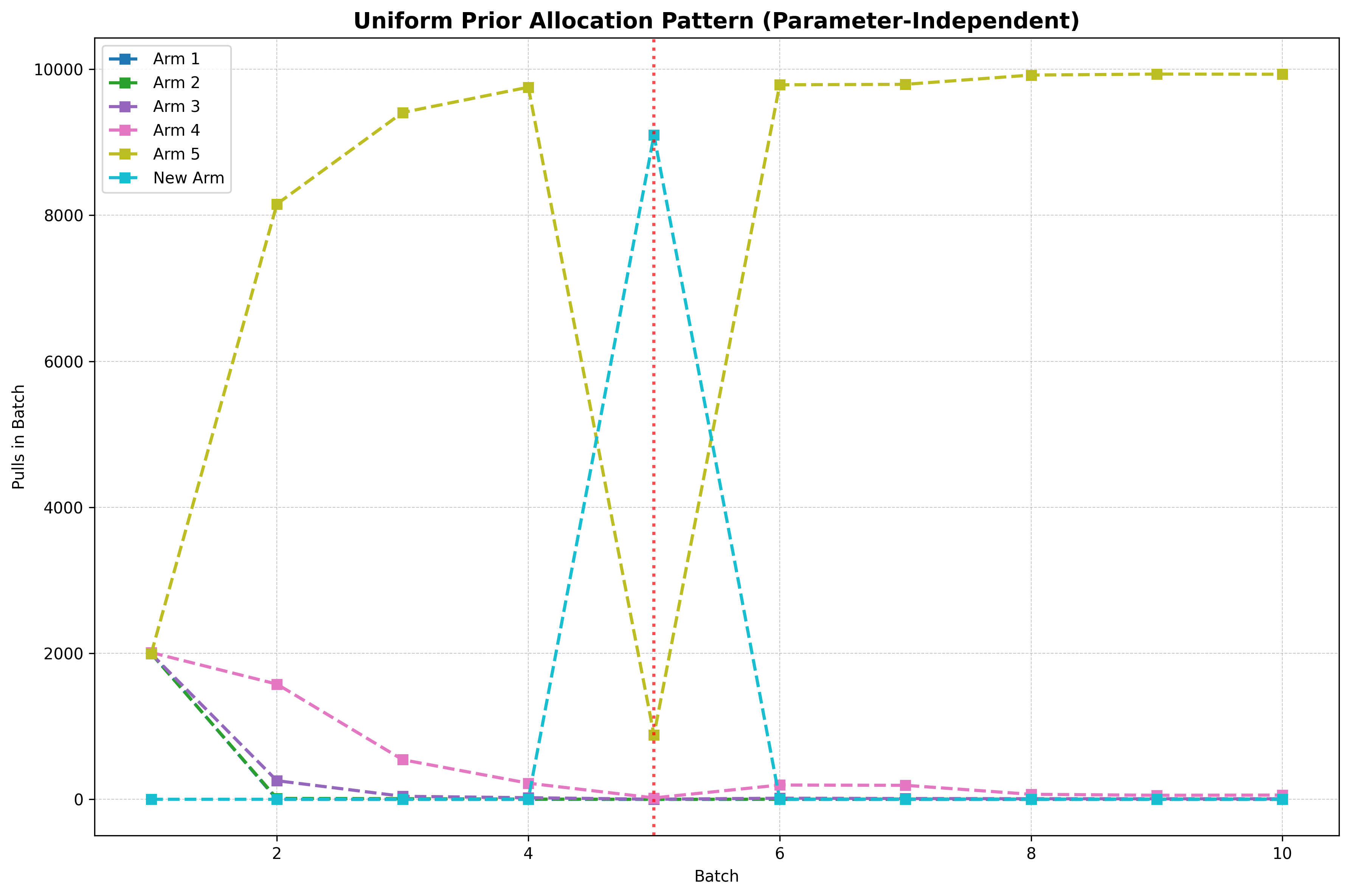}
\caption{(Simulation study) Uniform-prior allocation pattern showing persistent over-allocation to the weak new arm due to optimistic $\mathrm{Beta}(1,1)$ assumptions. This illustrates the mismatch between uniform priors and low-success-rate operational environments that motivates dynamic priors.}
\label{fig:uniform_allocation_pattern}
\end{figure}

\begin{figure*}[htbp]
\centering
\includegraphics[width=\textwidth]{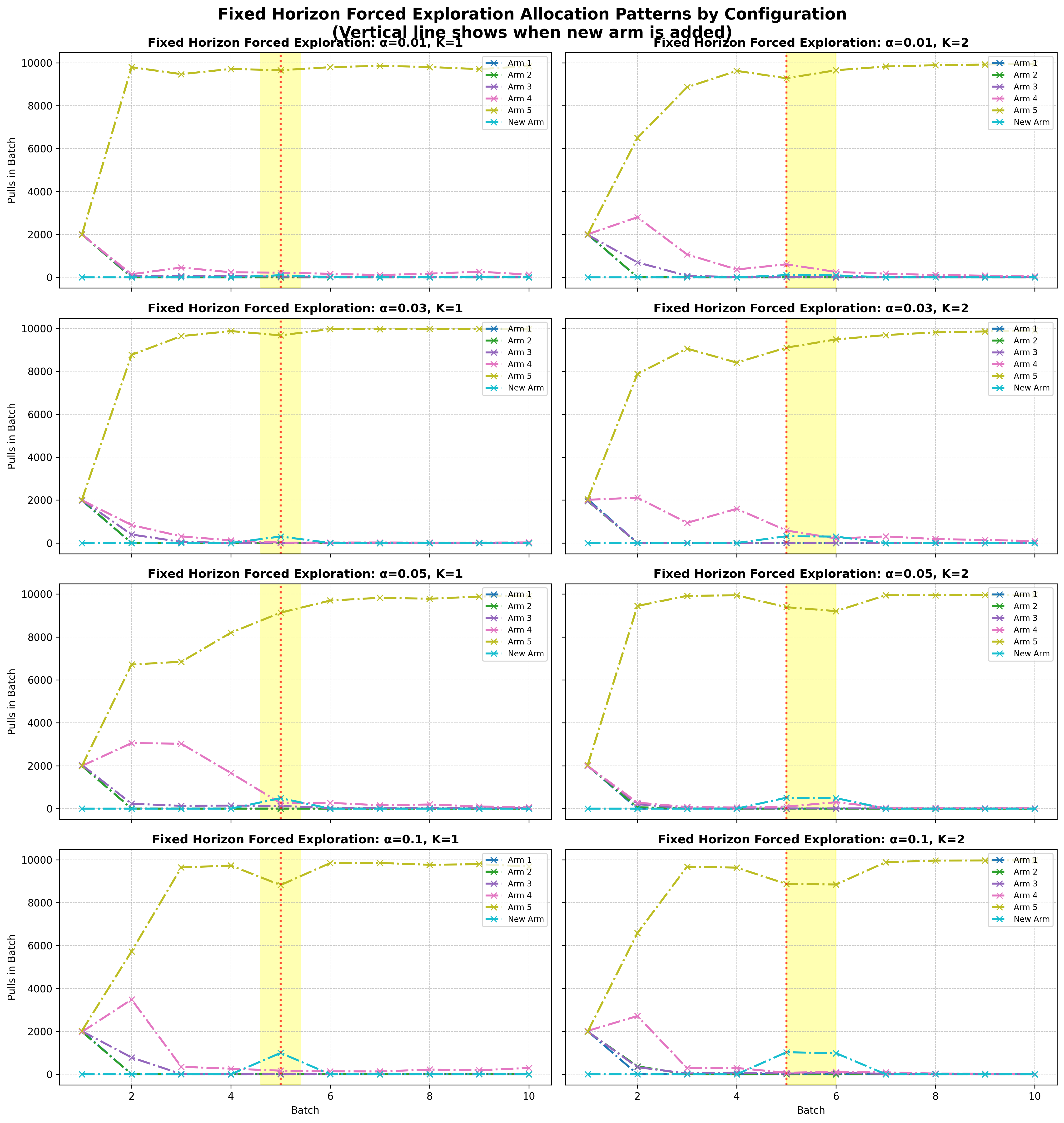}
\caption{(Simulation study) Allocation patterns for fixed-horizon forced exploration across parameter combinations. Yellow regions indicate forced exploration periods where the new arm receives guaranteed traffic. The resulting sharp phase transitions contrast with the smooth, posterior-driven adaptation of Dynamic Prior.}
\label{fig:forced_exploration_allocation_patterns}
\end{figure*}

The allocation patterns revealed in Figures~\ref{fig:dynamic_allocation_patterns}--\ref{fig:forced_exploration_allocation_patterns} provide insights into the fundamental differences between exploration strategies:

\textbf{Dynamic Prior Behavior}: Figure~\ref{fig:dynamic_allocation_patterns} demonstrates controlled exploration that quickly diminishes as evidence accumulates, achieving optimal exploration--exploitation balance. The method shows adaptive behavior where exploration intensity naturally decreases as the poor performance of the new arm becomes evident.

\textbf{Uniform Prior Limitations}: Figure~\ref{fig:uniform_allocation_pattern} reveals persistent over-allocation to the poor new arm due to optimistic $\mathrm{Beta}(1,1)$ prior assumptions that fail to match the low-success-rate operational environment.

\textbf{Forced Exploration Patterns}: Figure~\ref{fig:forced_exploration_allocation_patterns} shows sharp temporal boundaries between exploration and exploitation phases, contrasting with the smooth adaptive behavior of our dynamic prior approach.

\begin{figure*}[htbp]
\centering
\includegraphics[width=1.0\textwidth]{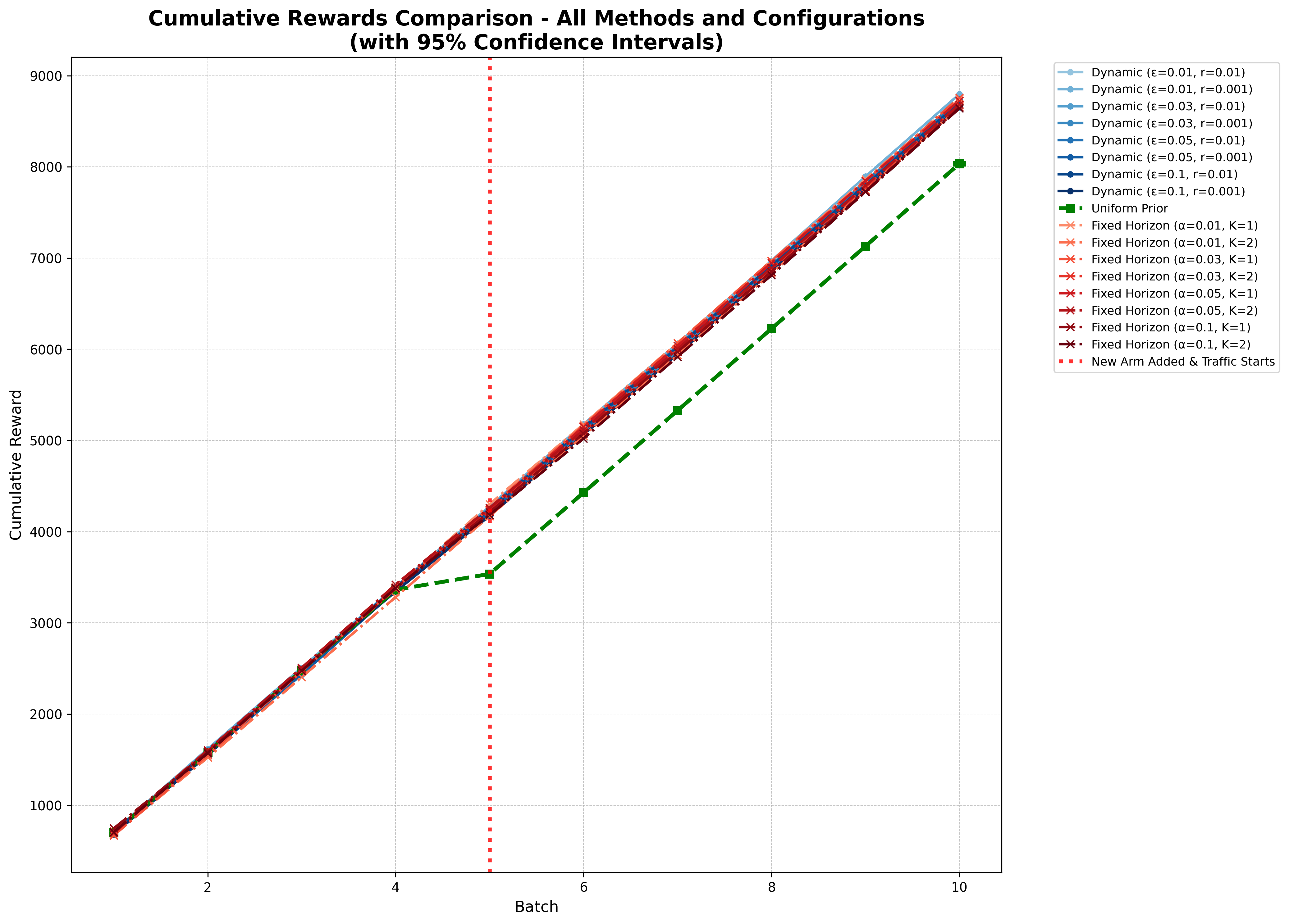}
\caption{(Simulation study) Cumulative reward trajectories for all methods and configurations. The vertical dashed line at batch 5 indicates the introduction of the new arm. Dynamic Prior configurations consistently outperform baselines, with error bars showing 95\% confidence intervals for final rewards.}
\label{fig:cumulative_rewards_comparison}
\end{figure*}

\section{Online Evaluation}

To validate the Dynamic Prior framework, we conducted a large-scale online experiment on a global gaming platform over a 20-day duration. The evaluation focuses on the ``Mid-flight Insertion'' use case, where a new thumbnail (serving as a new arm) is added to an active bandit instance without interrupting the service of existing arms.

\subsection{Experimental Design and Causal Validity}
Unlike standard user-randomized A/B tests, this experiment employed \textbf{game-level randomization} to prevent data leakage, as exploration training data are shared across treatment arms within a single game context. 

\begin{itemize}
    \item \textbf{Control Group ($\mathcal{C}$):} Current production logic. Upon the introduction of a new thumbnail, the algorithm undergoes a hard reset, discarding all posterior historical data for existing arms and re-initializing with $\mathrm{Beta}(1,1)$ priors.
    \item \textbf{Treatment Group 1 ($\mathcal{T}_1$):} Dynamic Prior with conservative exploration ($\varepsilon = 3\%$) and prior strength ($r = 0.1$).
    \item \textbf{Treatment Group 2 ($\mathcal{T}_2$):} Dynamic Prior with moderate exploration ($\varepsilon = 10\%$) and prior strength ($r = 0.1$).
\end{itemize}

The choice of prior strength $r = 0.1$ is strategically calibrated to balance prior stability with posterior responsiveness. Since the algorithm operates on a 7-day (168-hour) sliding window of historical data, the effective sample size $n_k \times r$ represents approximately $16.8$ hours of observations. This strength provides a robust buffer that prevents the prior from being prematurely overwritten by short-term noise---critical given the $3$--$7$ hour data SLA delays in our production pipeline---while ensuring that the model can still adapt and update within a single day as new evidence accumulates. Empirical observations in our later analysis confirm this balance: winning arms typically secure majority traffic within $24$ hours, while suboptimal arms are suppressed within a similar timeframe.

To ensure causal validity under game-level assignment, we utilized a \textbf{Difference-in-Differences (DiD)} regression weighted by game impressions, controlling for game-specific and temporal confounders. The primary success metric was \textbf{Qualified Play-Through Rate (QPTR)}, defined as a play-through on an impression if the session duration exceeds a predefined time threshold. Pre-experiment A/A testing confirmed no significant bias across arms ($p > 0.05$).

\subsection{Primary Results: Engagement and Efficiency}

Both treatment arms exhibited statistically significant lifts in QPTR, validating the hypothesis that maintaining posterior continuity reduces the cost of exploration. Table~\ref{tab:results} summarizes the key performance indicators.

\begin{table}[h]
\centering
\caption{(Online experiment) QPTR lift and convergence metrics for mid-flight insertion.}
\label{tab:results}
\begin{tabular}{lcccc}
\toprule
Arm & QPTR Lift & $p$-value & Regretted Imps. & Convergence \\
\midrule
$\mathcal{C}$ & -- & -- & 64.1\% & Baseline \\
$\mathcal{T}_1$ & +0.20\% & 0.034 & 52.1\% & Slow \\
$\mathcal{T}_2$ & +0.19\% & 0.016 & 50.6\% & Fast \\
\bottomrule
\end{tabular}
\end{table}

\textbf{Exploration Efficiency:} A critical metric in our analysis is \textit{Regretted Impressions}---the percentage of impressions allocated to non-winning arms, calculated retrospectively post-convergence. $\mathcal{T}_2$ achieved a \textbf{21.0\% relative reduction} in regretted impressions compared to the Control ($50.6\%$ vs. $64.1\%$). This efficiency stems from the model's ability to leverage prior knowledge; if an existing arm is already a strong winner, the algorithm avoids the wasteful ``re-discovery'' phase necessitated by a production reset.

\subsection{Deep Dive: Exploration Dynamics}

To select the optimal hyperparameter $\varepsilon$, we analyzed the trade-off between statistical efficiency and ecosystem health.

\textbf{1. New Thumbnail Breakout Rate:} We measured the frequency with which a newly introduced thumbnail successfully became the winning arm. $\mathcal{T}_1$ ($71.65\%$) and $\mathcal{T}_2$ ($71.73\%$) showed no statistically significant difference ($p = 0.984$), suggesting that an initial $\varepsilon$ as low as $3\%$ is sufficient to identify high-performing content.

\textbf{2. Developer Perception:} As a developer-facing platform, creator satisfaction is a critical objective. Qualitative analysis suggests that the initial $3\%$ exploration rate in $\mathcal{T}_1$ can be perceived by developers as ``under-exploration,'' leading to frustration when viewing bandit traffic dashboards. This is primarily because a $3\%$ allocation is barely discernible in the dashboard graph, whereas a $10\%$ allocation provides a clearly visible signal of exploration. As illustrated in Figures~\ref{fig:g2_win} and \ref{fig:g2_lose}, $\mathcal{T}_2$ aligns more closely with developer expectations by providing a more substantial and observable ``launch'' window for new creative assets.

\subsection{Production Launch}

\begin{figure*}[htbp]
\centering
\includegraphics[width=\textwidth]{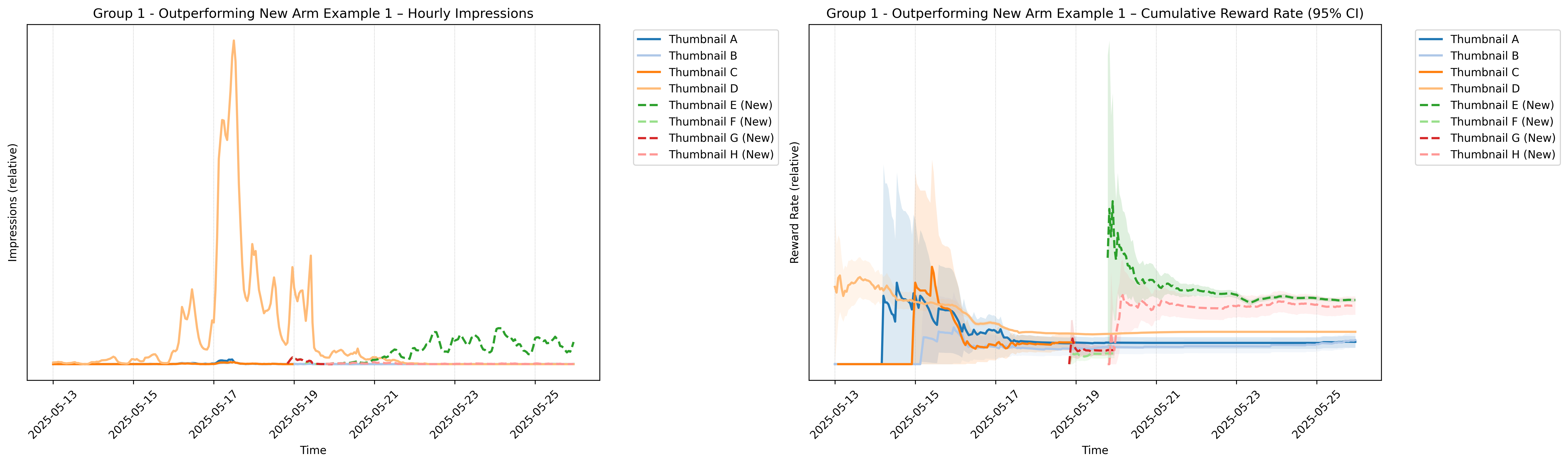}
\caption{(Online experiment; example $\mathcal{T}_1$) New arm eventually wins. The newly added thumbnail (dashed line) receives controlled exploration at $\varepsilon=3\%$ and then gains share as evidence accumulates, ultimately becoming the dominant choice.}
\label{fig:g1_win}
\end{figure*}

\begin{figure*}[htbp]
\centering
\includegraphics[width=\textwidth]{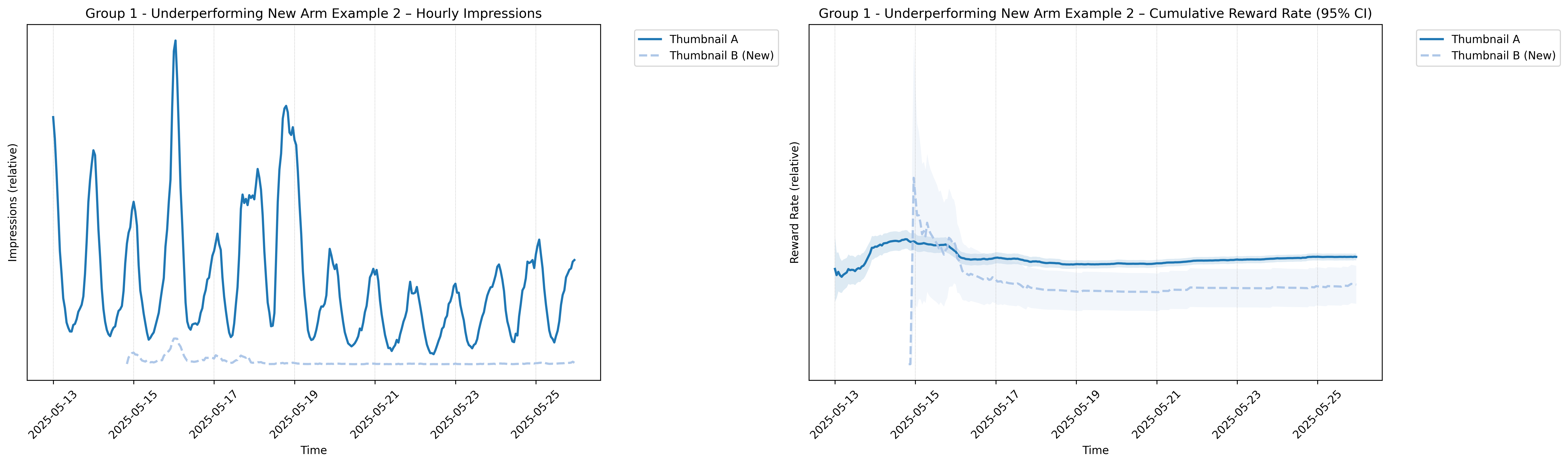}
\caption{(Online experiment; example $\mathcal{T}_1$) New arm underperforms. The new thumbnail (dashed line) receives initial exploration traffic but rapidly loses allocation as evidence reveals poor performance relative to existing options.}
\label{fig:g1_lose}
\end{figure*}

\begin{figure*}[htbp]
\centering
\includegraphics[width=\textwidth]{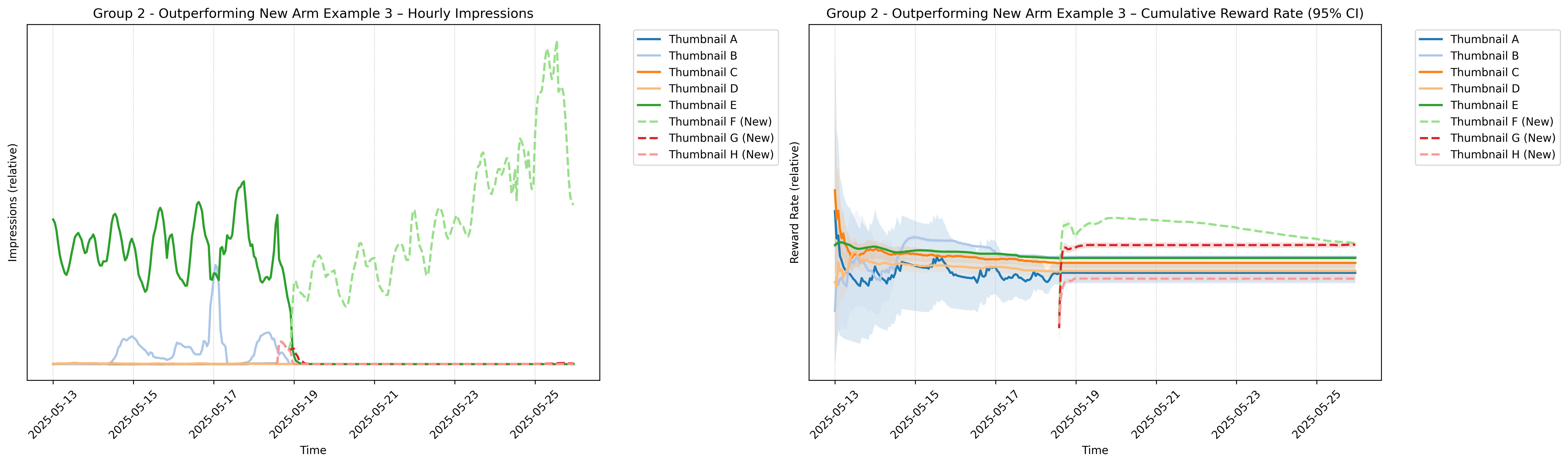}
\caption{(Online experiment; example $\mathcal{T}_2$) Faster discovery under higher exploration. Compared to $\mathcal{T}_1$, the higher $\varepsilon=10\%$ allocates more initial traffic to the new arm, accelerating identification of superior performance.}
\label{fig:g2_win}
\end{figure*}

\begin{figure*}[htbp]
\centering
\includegraphics[width=\textwidth]{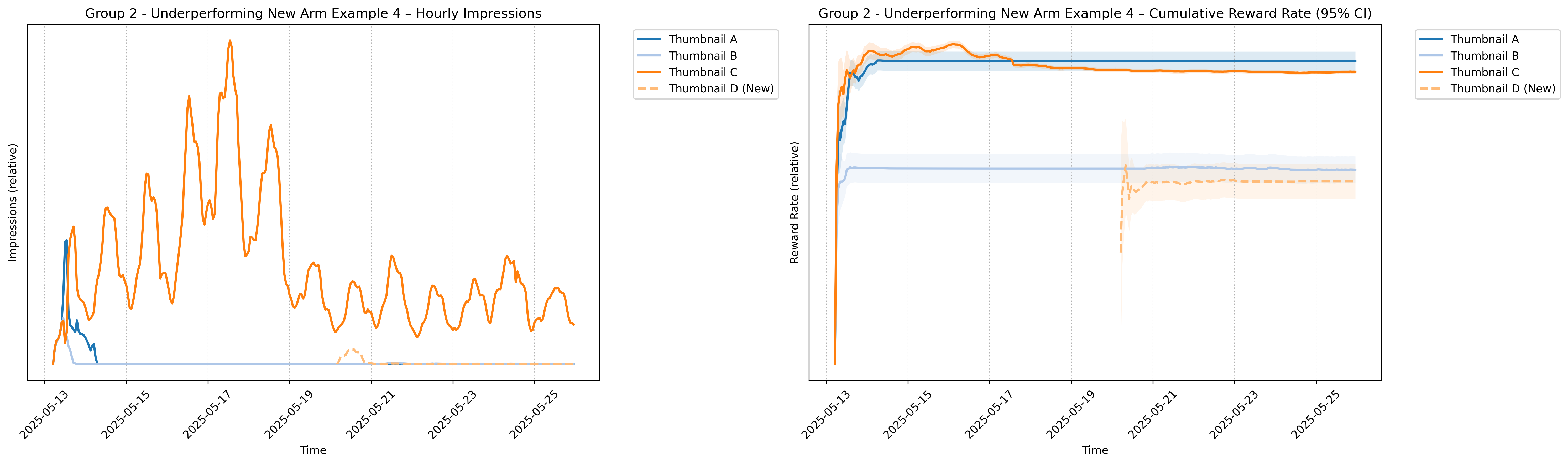}
\caption{(Online experiment; example $\mathcal{T}_2$) Robust suppression of a weak new arm. Despite the higher 10\% initial allocation, the system quickly reduces traffic once poor performance is observed, indicating strong regret control under more aggressive exploration.}
\label{fig:g2_lose}
\end{figure*}

Based on the superior convergence speed in complex scenarios and higher developer satisfaction scores, we launched \textbf{$\mathcal{T}_2$ ($\varepsilon = 10\%, r = 0.1$)} to production. This configuration maintains the engagement lift of the Dynamic Prior while ensuring the algorithm remains robust and transparent to the creator community. Figures~\ref{fig:g1_win} through \ref{fig:g2_lose} demonstrate how impression allocation adapts dynamically based on empirical performance across different treatment levels.

\section{Conclusion and Future Directions}

In this paper, we addressed the critical challenge of cold-start exploration in large-scale recommender systems, particularly under the constraints of batched production updates and delayed feedback. We introduced a \textit{Dynamic Prior} framework for Thompson Sampling that replaces uninformed uniform priors with an adaptive Bayesian approach. By deriving a closed-form quadratic solution to ensure that new arms compete with established winners at a pre-specified target probability $\varepsilon$, we provided the first intrinsic mechanism for exact exploration probability control within the Thompson Sampling framework.

Our theoretical validation and offline simulations demonstrated that the dynamic prior achieves high-precision control across diverse operational conditions. Furthermore, our large-scale online evaluation on a production thumbnail personalization system serving millions of users confirmed the practical efficacy of our approach. The transition to the dynamic prior resulted in a statistically significant $+0.19\%$ lift in Qualified Play-Through Rate (QPTR) and a $21.0\%$ relative reduction in regretted impressions. 

Beyond quantitative gains, our study highlighted the importance of stakeholder alignment in platform ecosystems. This algorithm provides the feasibility for changing the exploration rate to balance algorithmic efficiency with developer observability, ensuring that the discovery process remains transparent and satisfactory for content creators.

Future research will explore the integration of this methodology with contextual bandit architectures to leverage high-dimensional side information while maintaining the exact exploration guarantees established in this work. Finally, we plan to investigate multi-objective dynamic priors that can simultaneously optimize for engagement, fairness, and long-term ecosystem diversity.

\bibliographystyle{ACM-Reference-Format}
\bibliography{reference}

\appendix
\section{Notation Summary}

\textbf{Key Notation:}
\begin{itemize}
\item $i, j, k$: Arm indices (generic arm $i$, new arm $j$, best-performing arm $k$)
\item $\Theta_i$: Random variable for the true success rate of arm $i$
\item $\theta_i^{(t)}$: Sample drawn from the posterior of $\Theta_i$ at round $t$
\item $\alpha_i, \beta_i$: Parameters of the Beta posterior for arm $i$
\item $n_i$: Number of observations for arm $i$
\item $\hat{p}_i$: Observed success rate for arm $i$
\item $r$: Prior strength factor relative to best arm's experience
\item $\varepsilon$: Target exploration probability for a new arm
\item $z_{\varepsilon}$: Normal quantile $\Phi^{-1}(\varepsilon)$
\item $T_{z\varepsilon}$: Square of the normal quantile $z_{\varepsilon}^2$
\item $q_j$: Derived prior mean success rate for new arm $j$
\item $n_k$: Number of observations for the current best arm $k$
\item $\hat{p}_k$: Observed success rate for the current best arm $k$
\item $X_j$: Random variable for success rate of new arm $j$
\item $Y_k$: Random variable for success rate of best arm $k$
\end{itemize}

\end{document}